%% file: main.tex
\documentclass[a4paper,UKenglish,cleveref,autoref]{lipics-v2021}

\usepackage{multirow}
\usepackage{makecell}
\usepackage{amsmath}
\usepackage{xcolor}
\usepackage[skins,most]{tcolorbox}
\usepackage{colortbl}
\usepackage[linesnumbered,ruled,vlined]{algorithm2e}
\usepackage{subcaption,siunitx,booktabs}
\usepackage{placeins}
\usepackage{pifont}

\newenvironment{finding}{\begin{tcolorbox}[enhanced,width=\linewidth,size=fbox,
colback=gray!10,colframe=gray!45,sharp corners,boxrule=0.5pt]}{\end{tcolorbox}}

\title{SemiScope: Disentangling Classifier Tuning and Joint Optimization in Semi-Supervised Security Classification}
\titlerunning{SemiScope: Disentangling Classifier Tuning and Joint Optimization}

\author{Rui Shu}{North Carolina State University}{}{}{}
\author{Tianpei Xia}{North Carolina State University}{}{}{}
\author{Jingzhu He}{ShanghaiTech University}{}{}{}

\authorrunning{Rui Shu, Tianpei Xia and Jingzhu He}

\Copyright{Rui Shu, Tianpei Xia and Jingzhu He}



\category{}
\relatedversion{}
\nolinenumbers

\begin{document}

\maketitle

\begin{abstract}
\noindent\textbf{Background.} Labeled data for security classification is scarce.
Semi-supervised learning (SSL) reduces this burden by propagating labels from a small labeled pool to larger unlabeled pools.
Yet security applications often use SSL as a black box: default parameters, a fixed classifier, and no explicit handling of pseudo-label-induced class imbalance.

\noindent\textbf{Aims.} Recent work reports sizeable gains from optimizing SSL pipelines through joint hyperparameter search, AutoML on pseudo-labeled data, or per-component tuning.
These gains are hard to attribute.
They may reflect useful joint SSL--classifier interactions, or they may mostly come from the simpler step of tuning the downstream classifier.
We disentangle these effects for binary tabular security classification with classical SSL and tree-based classifiers.

\noindent\textbf{Method.} We build SemiScope as an analysis instrument, not as a deployment recommendation.
It uses Bayesian Optimization to jointly choose SSL settings, confidence filtering, oversampling, classifier family, and classifier hyperparameters.
The key control is Tuned-Clf: it fixes SSL to defaults but receives the same 100-trial classifier search budget and the same validation-set decision-threshold tuning procedure as SemiScope.
At 10\% labels, we compare SemiScope and Tuned-Clf with paired TOST using a $\pm 1.0$ g-measure smallest effect size of interest.

\noindent\textbf{Results.} SemiScope exceeds every default SSL baseline on all five datasets, improving over the strongest default baseline by 0.7--12.7 g-measure points, depending on the dataset.
Under the equal-budget control, however, Tuned-Clf is statistically equivalent to the full joint pipeline on 4 out of 5 datasets, while Phishing is inconclusive.
Classifier hyperparameter optimization (HPO) alone recovers a median 86\% of SemiScope's gain over Default Self-Training (ST) + Random Forest (RF).
With equal budgets and symmetric threshold tuning, little extra value remains.

\noindent\textbf{Conclusions.} For these benchmarks, the reusable contribution is the decomposition protocol.
Practically, a simpler recipe is sufficient: use Self-Training, tune the classifier with Bayesian Optimization, and tune the decision threshold on validation data.
This recipe reaches within 1~g-measure of the Supervised RF reference at 20--30\% labels on four datasets and at 40\% labels on Drebin,
at the same or lower label rate than Default~ST~+~RF in every dataset.
\end{abstract}

\input{sections/introduction}
\input{sections/background}
\input{sections/methodology}
\input{sections/experiment}
\input{sections/results}
\input{sections/threats}
\input{sections/conclusion}

\section*{Data Availability}
\textbf{Data Availability:} The dataset supporting this study, including the mining scripts, raw repository data, and analysis notebooks, is openly available on Zenodo at \url{https://doi.org/10.5281/zenodo.21081753} under a CC-BY 4.0 license. The replication package contains all materials necessary to reproduce our findings.
\bibliography{main}

\end{document}

%% file: sections/introduction.tex
\section{Introduction}

Security practitioners increasingly rely on machine learning to detect threats such as spam,
malware~\cite{souri2018state}, and network intrusions~\cite{resende2018survey}.
The bottleneck is reliable labels: security labeling requires domain expertise, manual investigation,
and threat taxonomy judgment~\cite{tu2020better}.
Meanwhile, organizations collect large volumes of unlabeled security data, including network flows, URLs, and app manifests~\cite{sharafaldin2018toward,mamun2016detecting,arp2014drebin}.

Semi-supervised learning (SSL) addresses this bottleneck by propagating labels from a small labeled pool to a large unlabeled pool~\cite{zhu2009introduction}.
SSL has been applied to insider threat detection~\cite{le2019machine},
malware classification~\cite{kolosnjaji2016adaptive}, fraud detection~\cite{wang2021collaboration},
and intrusion detection~\cite{pallaprolu2016label}.
These applications often treat SSL as a black box preprocessing step.
They use default SSL hyperparameters, a single fixed downstream classifier, and no explicit mechanism for the class imbalance introduced by pseudo-labeling.
This is especially risky on imbalanced security data: with few labels, pseudo-labeling can further underrepresent the minority class, and naive oversampling can replicate those mistakes.

A semi-supervised security system is therefore a \textit{configurable pipeline}, not just an SSL algorithm.
Its SSL method, SSL hyperparameters, confidence filter, oversampling strategy, classifier family, classifier hyperparameters, and decision threshold can all change the final model.
Prior SSL studies in security~\cite{le2019machine,kolosnjaji2016adaptive,wang2021collaboration,pallaprolu2016label} report gains from tuning or automating parts of this pipeline, but they often compare tuned pipelines against default classifiers or use asymmetric threshold evaluation.
Such protocols show that tuning helps, but they do not identify which part of the pipeline caused the gain.
They leave open which components account for most of the observed improvement,
whether joint SSL--classifier optimization outperforms classifier tuning at equal compute budget,
and whether unequal search budgets or different threshold policies inflate the apparent gains.
We answer with baselines that use equal budgets, symmetric threshold tuning, and fixed rules.

The Phishing result illustrates the confound (Table~\ref{tbl:rq1}).
At 10\% labels, a full joint search reaches 96.1~g-measure, far above default label propagation (LP)~\cite{zhu2005semi} (87.7) and LP with Synthetic Minority Oversampling Technique (SMOTE)~\cite{chawla2002smote} (89.5).
However, tuning only the downstream classifier on the same default LP output reaches 95.5 under the same budget of 100 trials.
In this example, the classifier tuning control explains almost all of the apparent advantage of joint pipeline optimization.
This motivates our central question: \textit{how much of the gain comes from joint optimization, and how much from classifier tuning alone?}

To answer, we construct \textbf{SemiScope} as a controlled joint search instrument.
SemiScope uses Bayesian optimization~\cite{snoek2012practical} over SSL settings, classifier family and hyperparameters, confidence filtering, and oversampling with target ratios.
We do not present SemiScope as the method practitioners should deploy by default.
Instead, it provides a strong joint search reference under a fixed budget of 100 trials.
Our key comparison is Tuned-Clf, a control with the same budget that keeps SSL at defaults and spends the same 100 trials on the same classifier family and hyperparameter space.
All treatments tune the decision threshold on validation data before final test evaluation.
This design isolates classifier hyperparameter optimization (HPO) from the residual value of the joint formulation.
To our knowledge, no prior SSL study in security has made this decomposition with symmetric thresholds and equal budgets.

The decomposition yields three findings:
\textbf{(F1)} SemiScope beats every default baseline on all five datasets, consistent with prior reports.
The gains over the strongest default baseline range from $+$0.7 to $+$12.7~g-measure.
\textbf{(F2)} Against the classifier tuner with the same budget, the full joint pipeline is statistically equivalent on four of five datasets under paired TOST~\cite{lakens2017equivalence} with SESOI~$=\pm 1.0$, while the Phishing dataset is inconclusive.
Classifier hyperparameter optimization (HPO) alone recovers a \textit{median 86\%} of SemiScope's gain over default Self-Training (ST)~\cite{scudder1965probability,lee2013pseudo} + Random Forest (RF)~\cite{breiman2001random}.
\textbf{(F3)} Decision threshold tuning materially affects the comparison.
Almost all tuned thresholds fall below the default 0.5, so evaluation with a fixed threshold can collapse minority recall and distort comparisons.

Thus, the main contribution is not a new optimizer, but an empirical decomposition of where SSL pipeline gains come from.
Under fair controls, gains from classifier tuning can be mistaken for gains from joint SSL pipeline optimization.
The appropriate benchmark is therefore not a joint optimizer versus default SSL, but a joint optimizer versus a classifier tuner with the same budget and threshold policy.
We make three contributions:
\begin{itemize}
    \item \textbf{Controlled decomposition.} SemiScope serves as the joint search \textit{instrument}, while Tuned-Clf serves as the classifier tuning \textit{control} with the same budget.
    This separates gains due to joint SSL--classifier optimization from gains due to classifier HPO.
    \item \textbf{Symmetric threshold protocol.} Each treatment tunes its decision threshold on validation data before test evaluation.
    This avoids comparing methods under different threshold assumptions on datasets with 12--20\% minority classes.
    \item \textbf{Actionable recipe.} Use Self-Training as default SSL, add classifier HPO with Bayesian optimization, and tune the validation threshold.
    This reaches within 1~g-measure of the Supervised RF reference at 20--30\% labels on four datasets and at 40\% labels on Drebin,
    at the same or lower label rate than Default~ST~+~RF in every dataset.
\end{itemize}

%% file: sections/background.tex
\section{Background and Related Work}\label{sec:background}

\subsection{Semi-Supervised Learning in Security}

Semi-supervised learning (SSL)~\cite{zhu2009introduction} trains models from a small labeled pool and a larger unlabeled pool.
This setting matches many security tasks, where unlabeled data are abundant but reliable labels are quite expensive.
A recent survey~\cite{mvula2024survey} catalogs growing SSL use in cybersecurity, including insider threat detection~\cite{le2019machine},
malware classification~\cite{alabdulmohsin2016content,kolosnjaji2016adaptive},
fraud detection~\cite{wang2021collaboration}, and intrusion detection~\cite{pallaprolu2016label}.

The classical SSL methods used in this paper fall into two families~\cite{van2020survey}.
\textit{Graph methods}, such as label propagation~\cite{zhu2005semi} and label spreading~\cite{zhou2003learning},
propagate labels along similarity graph edges.
\textit{Wrapper methods}, such as self-training~\cite{scudder1965probability,lee2013pseudo},
iteratively expand the labeled pool using a classifier's own predictions.
Both families expose the downstream classifier to pseudo-labels or class scores on unlabeled samples.
That coupling is important for imbalanced security data: ambiguous samples tend to be assigned to the majority class, so pseudo-labeling can \textit{worsen} class imbalance~\cite{iscen2019label}.
Imbalance handling is therefore part of the SSL pipeline rather than a separate step after SSL.

\subsection{Oversampling for Class Imbalance}

Oversampling is a common response to class imbalance after the training labels have been fixed.
SMOTE~\cite{chawla2002smote} generates synthetic minority samples by interpolating between nearest neighbors,
Borderline-SMOTE~\cite{han2005borderline} focuses on boundary regions, and SMOTUNED~\cite{agrawal2018better} tunes SMOTE's $k$, $m$, $r_M$ via differential evolution.
In an SSL pipeline, however, oversampling is coupled to pseudo-label quality.
If pseudo-labels are wrong, synthetic samples can be generated in the wrong region and amplify the error.
For this reason, SemiScope uses a SMOTE strategy with a target class ratio rather than a fixed synthetic count.
The strategy is tuned jointly with SSL and classifier parameters, so the optimizer decides both oversampling activation and target ratio.

\subsection{Hyperparameter Optimization and Confidence Filtering}

A semi-supervised classification system has the same structure as a configurable software system: multiple choices interact, and exhaustive search is impractical~\cite{nair2018finding,jamshidi2017transfer}.
This motivates hyperparameter optimization (HPO) over pipeline choices, but it does not by itself show that the \textit{joint} search space is necessary.
SemiScope operationalizes joint search with Tree-structured Parzen Estimator (TPE)~\cite{bergstra2011algorithms} Bayesian optimization~\cite{snoek2012practical} via Optuna~\cite{akiba2019optuna}.
TPE handles the hierarchical conditional space where parameters for a method enter only when that method is sampled.
AutoML systems such as Auto-sklearn~\cite{feurer2015efficient},
TPOT~\cite{olson2016tpot}, and FLAML~\cite{wang2021flaml} automate classifier selection and tuning but operate on fully labeled data.
When they are applied after SSL, the pseudo-labels become fixed training labels for the AutoML system.
SemiScope instead keeps SSL algorithm choice, pseudo-label filtering, and oversampling inside the search space, rather than tuning only the downstream classifier.

Confidence filtering is one such choice specific to SSL, explored mainly in deep learning settings such as FixMatch~\cite{sohn2020fixmatch} and FreeMatch~\cite{wang2022freematch}.
FixMatch uses a fixed threshold, while FreeMatch introduces adaptive thresholding.
These methods target image data and rely on consistency loss from augmentations.
Their assumptions transfer poorly to tabular security data, where augmentation can be semantically ambiguous.
SemiScope adapts confidence filtering to \textit{tabular} security data with traditional ML classifiers by making the filter threshold a pipeline hyperparameter.
The optimizer can raise the threshold when pseudo-labels are noisy, or effectively disable filtering when it discards useful data.
Recent SSL and self-supervised methods for tabular data, including VIME~\cite{yoon2020vime}, SubTab~\cite{ucar2021subtab}, SCARF~\cite{bahri2021scarf}, and TabPFN~\cite{hollmann2022tabpfn}, target deep architectures and substantial compute.
Our study focuses instead on classical SSL with tree classifiers, a common setup in security SSL studies.

Testing joint search against independent tuning therefore depends on the comparison protocol, not only on the optimizer.
The software engineering literature on configurable systems shows that evaluation protocol can change method rankings.
Nair et al.~\cite{nair2018finding} demonstrated feature interactions that are not additive and break naive additive models.
Jamshidi et al.~\cite{jamshidi2017transfer} showed that configuration knowledge transfers across environments when the methodology controls for confounders.
At the SSL level, we therefore use a classifier tuning control with the same budget and identical threshold tuning for all treatments.

\begin{table}[!t]
\centering
\small
\setlength{\tabcolsep}{3pt}
\caption{Methodological coverage of representative SSL, AutoML, and security classification work.
\textbf{\checkmark}~supported; \ding{55}~not reported.
$^*$Joint classifier HPO, but no SSL tuning.
$^\dagger$Fixed SSL confidence threshold, and not tuned.}
\begin{tabular}{l|cccccc}
\hline
\textbf{Work} & \textbf{SSL Tune} & \textbf{Multi-Clf.} & \textbf{Imbal.} & \textbf{Thresh.} & \textbf{Multi-Seed} & \textbf{Joint HPO} \\ \hline \hline
Auto-sklearn~\cite{feurer2015efficient} & \ding{55} & \checkmark & \checkmark & \ding{55} & \ding{55} & \checkmark$^*$ \\
FLAML~\cite{wang2021flaml} & \ding{55} & \checkmark & \ding{55} & \ding{55} & \ding{55} & \checkmark$^*$ \\
FixMatch~\cite{sohn2020fixmatch} & \checkmark$^\dagger$ & \ding{55} & \ding{55} & \ding{55} & \ding{55} & \ding{55} \\
Le~\cite{le2019machine} & \ding{55} & \ding{55} & \ding{55} & \ding{55} & \ding{55} & \ding{55} \\
Wang~\cite{wang2021collaboration} & \ding{55} & \ding{55} & \ding{55} & \ding{55} & \ding{55} & \ding{55} \\
Kolosnjaji~\cite{kolosnjaji2016adaptive} & \ding{55} & \ding{55} & \ding{55} & \ding{55} & \ding{55} & \ding{55} \\
\textbf{SemiScope} & \textbf{\checkmark} & \textbf{\checkmark} & \textbf{\checkmark} & \textbf{\checkmark} & \textbf{\checkmark} & \textbf{\checkmark} \\
\hline
\end{tabular}
\label{tbl:positioning}
\end{table}

\subsection{Positioning}

Taken together, the SSL, imbalance, HPO, and confidence filtering literatures position our work as a benchmarking and decomposition study.
The study also follows evaluation guidance for security: Arp et al.~\cite{arp2022and} identify common pitfalls in ML-based security research, including unfair baselines and insufficiently diverse evaluation.
We follow this guidance by using identical decision threshold tuning across treatments and evaluating five datasets spanning four task types.
This matters for SSL pipelines because extra search effort or a favorable threshold can make a method appear better.
A tuned joint pipeline compared only with defaults therefore mixes SSL choices with classifier tuning.
Table~\ref{tbl:positioning} makes these controls explicit.
SemiScope addresses the gap by decomposing these elements under equal budgets, repeated seeds, and symmetric thresholds.

%% file: sections/methodology.tex
\section{SemiScope: Pipeline and Decomposition Framework}\label{sec:methodology}

\begin{figure}[!t]
\centering
\includegraphics[width=0.94\linewidth,keepaspectratio]{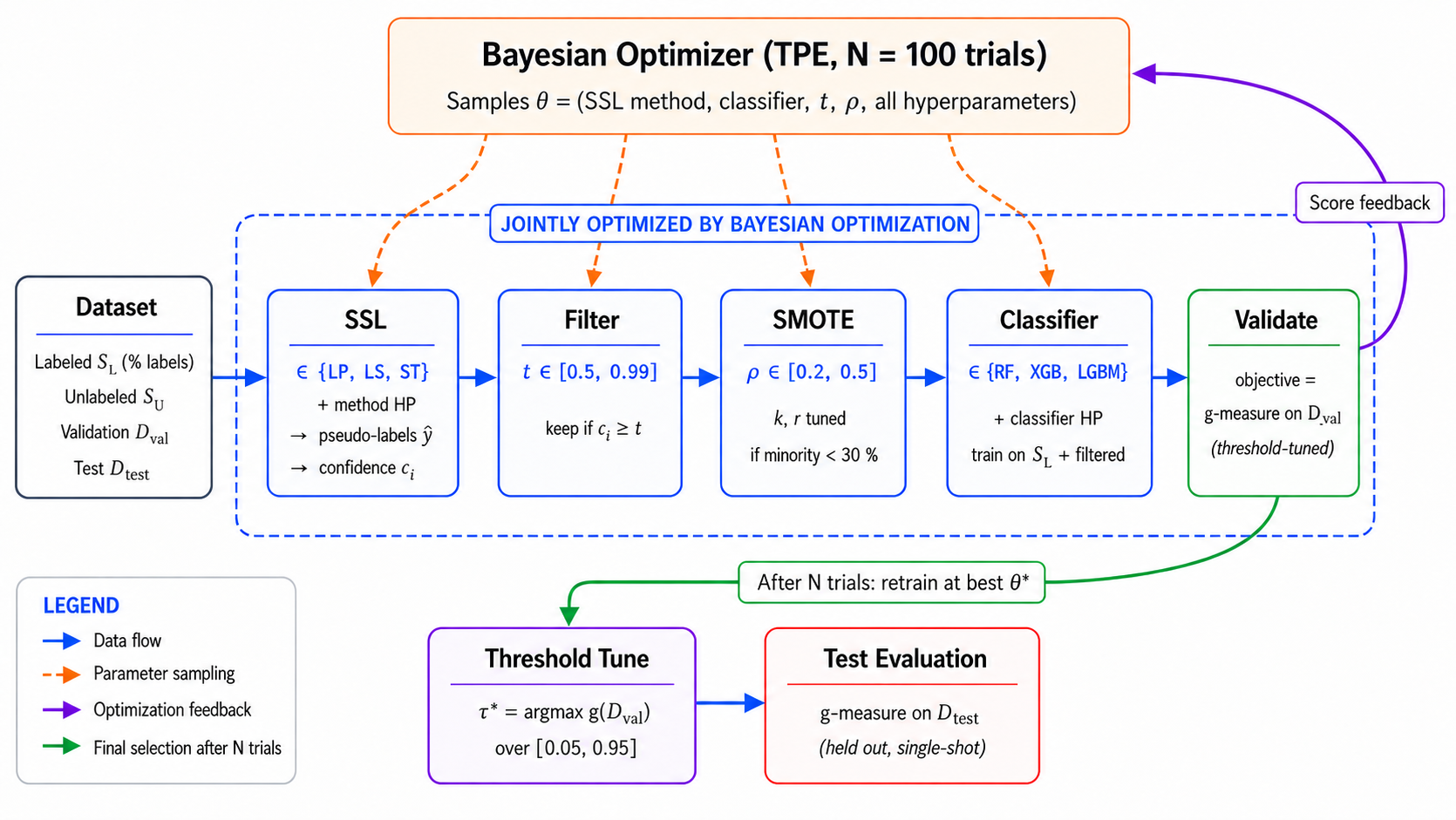}
\caption{SemiScope pipeline and shared evaluation protocol.
Bayesian optimization~\cite{snoek2012practical} with TPE~\cite{bergstra2011algorithms} (100 trials) samples pipeline parameters $\theta$ for SSL, filtering, oversampling, and classification.
Validation g-measure selects the best configuration $\theta^*$.
The final decision threshold is tuned on $D_{\text{val}}$, and every baseline uses the same threshold-tuning and held-out-test protocol.}
\label{fig:semiscope_pipeline}
\end{figure}

To make the decomposition operational, we first formalize SemiScope as a joint search over pipeline configurations.
Given a labeled set $S_L$, an unlabeled set $S_U$, and a validation set $D_{\text{val}}$,
SemiScope seeks the configuration $\theta^*$ that maximizes g-measure on $D_{\text{val}}$:
\begin{equation}
    \begin{aligned}
        \hat{f}_{\theta} &= \text{Pipeline}(S_L, S_U; \theta),\\
        \theta^* &= \operatorname*{arg\,max}_{\theta \in \Theta}
        \text{g-measure}\big(\hat{f}_{\theta}, D_{\text{val}}\big),
    \end{aligned}
\end{equation}
where $\theta$ denotes one complete pipeline configuration, $\hat{f}_{\theta}$ is the trained predictor returned by $\text{Pipeline}(\cdot; \theta)$, and $\Theta$ is the set of all configurations considered in the search.
The space spans SSL method, SSL hyperparameters,
confidence threshold, classifier type, classifier hyperparameters,
and SMOTE~\cite{chawla2002smote} parameters (Table~\ref{tbl:sslRange}).
Figure~\ref{fig:semiscope_pipeline} shows the corresponding stages: SSL pseudo-labeling, confidence filtering, oversampling, classification, and validation-set threshold tuning.
The reported numbers are computed once on held-out $D_{\text{test}}$ as described in Section~\ref{sec:framework}.
This formulation has two roles in the paper:
it defines the full joint-search reference, and it provides axes that can be held fixed in matched baselines.
In the rest of this section, we first define the optimized components, specify the shared evaluation protocol and search space, and then explain the controls used for decomposition.

\subsection{Semi-Supervised Learning Methods}\label{sec:ssl_methods}

The pipeline begins by choosing how pseudo-labels are assigned to $S_U$.
SemiScope searches over three classical SSL methods, each returning a pseudo-label $\hat{y}_i$ and confidence score $c_i$ for every unlabeled sample.
\textbf{Label Propagation (LP)}~\cite{zhu2005semi} constructs a similarity graph over all data points and propagates labels from labeled to unlabeled nodes via a Laplacian transition matrix, yielding a class distribution $p(c|x_i)$; its tunable hyperparameters are kernel type (KNN or RBF), kernel bandwidth, and maximum iterations.
\textbf{Label Spreading (LS)}~\cite{zhou2003learning} follows the same graph-based intuition, but uses the normalized graph Laplacian and a clamping factor $\alpha \in (0,1)$, where smaller $\alpha$ trusts the original labels more and larger $\alpha$ lets propagated mass flow more freely.
\textbf{Self-Training (ST)}~\cite{scudder1965probability,lee2013pseudo} is the wrapper alternative: it iteratively trains a base classifier, adds high-confidence predictions to the labeled pool, and retrains.
Its internal acceptance threshold is \texttt{st\_threshold} $\in [0.6, 0.99]$.
In SemiScope, the ST base estimator is coupled to the optimizer's final classifier choice; for example, the base estimator uses the same classifier family and hyperparameters sampled for the final stage.
This aligns SSL's decision boundary with the downstream classifier's.

Before these SSL outputs enter the next stage, we settle two notational details.
First, the per-sample confidence used by the external filter is $c_i = \max_c p(c|x_i)$, computed from the converged label distribution (LP/LS) or from the base classifier's \texttt{predict\_proba} output (ST).
These scores are not on identical scales across methods, but because the optimizer selects the method and the confidence threshold $t$ jointly, it learns a threshold suited to whichever method is active.
Second, ST's internal \texttt{st\_threshold} (which controls which samples ST accepts inside its retraining loop) is distinct from the external confidence threshold $t$ (which is applied once after SSL completes), and therefore both are independently optimized.

\subsection{Confidence-Based Pseudo-Label Filtering}\label{sec:confidence_filter}

After SSL produces pseudo-labels, the next decision is whether all of them should be trusted for classifier training.
Standard SSL algorithms assign pseudo-labels to \textit{all} unlabeled samples, including points in ambiguous feature-space regions where the assignment is close to random.
Keeping these low-confidence assignments can corrupt the downstream classifier, and oversampling can then amplify the error:
if SMOTE generates synthetic samples around mis-labeled pseudo-points, it creates minority instances in the wrong regions.
SemiScope therefore filters pseudo-labels by confidence:
\begin{equation}
    S_{\text{filtered}} = \{(x_i, \hat{y}_i) \mid c_i \geq t\}, \quad t \in [0.5, 0.99].
\end{equation}
This threshold controls the precision--coverage tradeoff.
For example, at $t = 0.5$, the filter is effectively disabled because $\max_c p(c|x) \geq 0.5$ by definition in binary classification, so all pseudo-labels survive.
At $t = 0.9$, only high-confidence pseudo-labels pass, yielding a smaller but cleaner training set.
SemiScope makes $t$ a \textit{search variable} rather than a fixed cutoff, allowing the optimizer to keep reliable SSL outputs and filter noisy pseudo-labels.

\subsection{Classifier Selection and Ratio-Targeted Oversampling}\label{sec:classifiers_smote}

The filtered pseudo-labeled set is then combined with $S_L$ and passed to a downstream classifier.
Prior work on SSL for security often uses a single downstream classifier.
SemiScope instead treats the classifier family as a hyperparameter.
It searches over three tree-based models commonly used for tabular security classification:
\textbf{Random Forest}~\cite{breiman2001random} for bagged decision trees,
\textbf{XGBoost}~\cite{chen2016xgboost} for regularized gradient boosting, and
\textbf{LightGBM}~\cite{ke2017lightgbm} for leaf-wise gradient boosting with histogram splits.
Per-family hyperparameters are listed in Table~\ref{tbl:sslRange}.
Because each family exposes different hyperparameters, this creates a conditional search space.
The Optuna~\cite{akiba2019optuna} TPE sampler handles that structure natively: only the active family's parameters enter the surrogate, which keeps the effective search budget concentrated.

Before classifier training, the post-filter mixed set also determines whether oversampling is needed.
SemiScope uses SMOTE with tunable $k$ (neighbors) and $r_M$ (Minkowski distance power) following SMOTUNED~\cite{agrawal2018better}.
Unlike prior work that uses a fixed synthetic count, SemiScope uses a \textit{ratio-targeted} strategy.
The optimizer specifies a target minority fraction $\rho \in [0.2, 0.5]$ (i.e., the desired share of minority samples in the post-SMOTE mixed set), and the synthetic count is
\begin{equation}
    m = \max\Big(0,\; \lfloor \tfrac{\rho \cdot n_{\text{maj}}}{1 - \rho} \rfloor - n_{\text{min}}\Big)
\end{equation}
where $n_{\text{maj}}, n_{\text{min}}$ are post-filter class counts.
SMOTE is activated only when the post-filter minority fraction falls below $t_{\text{imb}}{=}30\%$.
When active, it draws $m$ synthetic minority samples by linear interpolation between $k$ nearest same-class neighbors under Minkowski power $r_M$.
The raw minority fractions in our datasets range from 12--20\% (Table~\ref{tbl:dataset}).
Given this range, we set $t_{\text{imb}}{=}30\%$ as a fixed guard threshold so post-filter sets that remain strongly skewed are eligible for SMOTE.
This threshold only controls activation:
once active, the target fraction $\rho$ controls how much rebalancing is applied, from a mild adjustment to $\rho \to 0.5$.

\subsection{Joint Optimization and Evaluation Protocol}\label{sec:framework}

Algorithm~\ref{alg:semiscope} shows how the preceding components are sampled in a single optimization loop.
Each trial draws a complete configuration $\theta \in \Theta$, allowing the TPE surrogate to model cross-component interactions such as which classifier family pairs best with which SSL method and confidence threshold.
We implement this search with Optuna and run $N{=}100$ trials per configuration.
Because method-specific parameters are active only when their method is sampled, each trial optimizes 16--18 parameters rather than every parameter in $\Theta$.
The decomposition study tests whether these modeled interactions matter.

\begin{algorithm}[!t]
\footnotesize
    \SetKwInput{Input}{Input}
    \SetKwInput{Output}{Output}
    \Input{training data $D_{\text{train}}$, validation data $D_{\text{val}}$, label rate $r$, imbalance threshold $t_{\text{imb}}$, trials $N$}
    \Output{optimized model $\text{clf}_{\text{opt}}$}
    Split $D_{\text{train}}$ into $S_L$ ($r$\%) and $S_U$ ($1{-}r$\%)\;
    \For{$i \gets 1$ \KwTo $N$}{
        Sample SSL method, classifier, confidence threshold $t$, SMOTE parameters ($\rho$, $k$, $r_M$), \texttt{use\_scaling}, and method/classifier-specific hyperparameters\;
        \lIf{\texttt{use\_scaling}}{standardize $S_L$, $S_U$, $D_{\text{val}}$ with a scaler fit on $S_L$}
        \BlankLine
        Run SSL on $S_L \cup S_U$ to obtain pseudo-labels and confidences\;
        Keep samples with confidence $\geq t$; concatenate with $S_L$ $\to D_{\text{mixed}}$\;
        \lIf{minority fraction in $D_{\text{mixed}} < t_{\text{imb}}$}{apply SMOTE($\rho$, $k$, $r_M$)}
        Train classifier; tune decision threshold on $D_{\text{val}}$\;
        Record g-measure on $D_{\text{val}}$\;
    }
    Let $\theta^*$ be the configuration with highest g-measure\;
    Retrain pipeline at $\theta^*$; re-tune $\tau$ on $D_{\text{val}}$ over $[0.05, 0.95]$ (step $0.01$)\;
    \Return $\text{clf}_{\text{opt}}$\;
    \caption{SemiScope joint optimization loop.}
    \label{alg:semiscope}
\end{algorithm}

After $N$ trials, we select the configuration with the highest validation g-measure and retrain the corresponding pipeline.
We then tune its decision threshold on $D_{\text{val}}$ by grid search over $[0.05, 0.95]$ (step 0.01) and evaluate once on $D_{\text{test}}$.
The same procedure is applied to every treatment, with the labeled/unlabeled split fixed per seed.
This symmetric threshold tuning is part of the estimator being compared, not a reporting detail:
on imbalanced data, a fixed $\tau{=}0.5$ cutoff can penalize different methods by different amounts.
Without this protocol, the Tuned-Clf vs.\ SemiScope comparison would confound models with thresholds.

\subsection{Search Space and Feature Scaling}\label{sec:hp_space}

The decomposition controls use the same classifier and pipeline search space, so we make that space explicit before defining the controls.
Table~\ref{tbl:sslRange} gives the hierarchical search space used by SemiScope and the controls.
The top-level choices, SSL method and classifier type, activate method-specific parameter subsets in each trial.
A boolean \texttt{use\_scaling} parameter toggles \texttt{StandardScaler}~\cite{pedregosa2011scikit} before SSL; the scaler is fit on the labeled training data and then applied to all splits.
Scaling can affect graph-based SSL because its kernels are distance-sensitive, while it is typically inert for tree models.

\begin{table}[!t]
\centering
\scriptsize
\setlength{\tabcolsep}{2.5pt}
\renewcommand{\arraystretch}{1.00}
\caption{SemiScope hierarchical search space.
Conditional parameters ($\dagger$) are active only for the selected SSL method or classifier; each trial activates 16--18 listed parameters.}
\begin{tabular}{@{}c|c|c|p{0.38\linewidth}@{}}
\hline
\textbf{Module} & \textbf{Parameter} & \textbf{Values / range} & \multicolumn{1}{c}{\textbf{Description}} \\ \hline \hline
\multirow{2}{*}{Selection} & SSL method & \{LP, LS, ST\} & Semi-supervised learning algorithm \\
 & Classifier & \{RF, XGBoost, LightGBM\} & Classification algorithm \\ \hline
\cellcolor{gray!8}Confidence & \cellcolor{gray!8}threshold $t$ & \cellcolor{gray!8}$[0.5, 0.99]$ & \cellcolor{gray!8}Minimum confidence to keep a pseudo-label \\ \hline
\multirow{4}{*}{LP$^\dagger$} & kernel & \{knn, rbf\} & Kernel function \\
 & gamma & $(10, 30)$ & RBF kernel parameter \\
 & n\_neighbors & $(5, 15)$ & KNN kernel parameter \\
 & max\_iter & $(500, 1500)$ & Maximum iterations \\ \hline
\cellcolor{gray!8} & \cellcolor{gray!8}kernel & \cellcolor{gray!8}\{knn, rbf\} & \cellcolor{gray!8}Kernel function \\
\cellcolor{gray!8} & \cellcolor{gray!8}gamma & \cellcolor{gray!8}$(10, 30)$ & \cellcolor{gray!8}RBF kernel parameter \\
\cellcolor{gray!8}LS$^\dagger$ & \cellcolor{gray!8}n\_neighbors & \cellcolor{gray!8}$(5, 15)$ & \cellcolor{gray!8}KNN kernel parameter \\
\cellcolor{gray!8} & \cellcolor{gray!8}alpha & \cellcolor{gray!8}$(0.1, 0.9)$ & \cellcolor{gray!8}Clamping factor \\
\cellcolor{gray!8} & \cellcolor{gray!8}max\_iter & \cellcolor{gray!8}$(500, 1500)$ & \cellcolor{gray!8}Maximum iterations \\ \hline
\multirow{2}{*}{ST$^\dagger$} & threshold & $(0.6, 0.99)$ & Self-training confidence threshold \\
 & max\_iter & $(5, 50)$ & Maximum self-training iterations \\ \hline
\cellcolor{gray!8} & \cellcolor{gray!8}n\_estimators & \cellcolor{gray!8}$[50, 200]$ & \cellcolor{gray!8}Number of trees \\
\cellcolor{gray!8} & \cellcolor{gray!8}max\_depth & \cellcolor{gray!8}$[1, 25]$ & \cellcolor{gray!8}Maximum tree depth \\
\cellcolor{gray!8} & \cellcolor{gray!8}max\_features & \cellcolor{gray!8}\{sqrt, log2\} & \cellcolor{gray!8}Features for best split \\
\cellcolor{gray!8}RF$^\dagger$ & \cellcolor{gray!8}min\_samples\_leaf & \cellcolor{gray!8}$[1, 25]$ & \cellcolor{gray!8}Minimum samples at leaf \\
\cellcolor{gray!8} & \cellcolor{gray!8}min\_samples\_split & \cellcolor{gray!8}$[2, 25]$ & \cellcolor{gray!8}Minimum samples to split a node \\
\cellcolor{gray!8} & \cellcolor{gray!8}max\_leaf\_nodes & \cellcolor{gray!8}$[2, 100]$ & \cellcolor{gray!8}Maximum leaf nodes \\
\cellcolor{gray!8} & \cellcolor{gray!8}bootstrap & \cellcolor{gray!8}\{True, False\} & \cellcolor{gray!8}Bootstrap sampling \\ \hline
\multirow{7}{*}{XGBoost$^\dagger$} & n\_estimators & $[50, 300]$ & Number of boosting rounds \\
 & max\_depth & $[3, 10]$ & Maximum tree depth \\
 & learning\_rate & $[0.01, 0.3]$ & Step size shrinkage \\
 & subsample & $[0.6, 1.0]$ & Row subsampling ratio \\
 & colsample\_bytree & $[0.6, 1.0]$ & Column subsampling ratio \\
 & min\_child\_weight & $[1, 10]$ & Minimum instance-weight sum per child \\
 & gamma & $[0.0, 5.0]$ & Minimum loss reduction to split \\ \hline
\cellcolor{gray!8} & \cellcolor{gray!8}n\_estimators & \cellcolor{gray!8}$[50, 300]$ & \cellcolor{gray!8}Number of boosting rounds \\
\cellcolor{gray!8} & \cellcolor{gray!8}max\_depth & \cellcolor{gray!8}$[3, 15]$ & \cellcolor{gray!8}Maximum tree depth \\
\cellcolor{gray!8} & \cellcolor{gray!8}num\_leaves & \cellcolor{gray!8}$[15, 127]$ & \cellcolor{gray!8}Maximum leaves per tree \\
\cellcolor{gray!8}LightGBM$^\dagger$ & \cellcolor{gray!8}learning\_rate & \cellcolor{gray!8}$[0.01, 0.3]$ & \cellcolor{gray!8}Step size shrinkage \\
\cellcolor{gray!8} & \cellcolor{gray!8}subsample & \cellcolor{gray!8}$[0.6, 1.0]$ & \cellcolor{gray!8}Row subsampling ratio \\
\cellcolor{gray!8} & \cellcolor{gray!8}colsample\_bytree & \cellcolor{gray!8}$[0.6, 1.0]$ & \cellcolor{gray!8}Column subsampling ratio \\
\cellcolor{gray!8} & \cellcolor{gray!8}min\_child\_samples & \cellcolor{gray!8}$[5, 50]$ & \cellcolor{gray!8}Minimum data in a leaf \\ \hline
\multirow{3}{*}{SMOTE} & k & $[1, 20]$ & Number of neighbors \\
 & $r_M$ & $[1, 6]$ & Minkowski distance power \\
 & $\rho$ & $[0.2, 0.5]$ & Target minority fraction (post-SMOTE) \\ \hline
\cellcolor{gray!8}Scaling & \cellcolor{gray!8}use\_scaling & \cellcolor{gray!8}\{True, False\} & \cellcolor{gray!8}Apply StandardScaler before SSL \\ \hline
\end{tabular}
\label{tbl:sslRange}
\end{table}

\subsection{Decomposition Controls}\label{sec:decomposition}

The joint optimizer is a strong reference, but it cannot attribute gains by itself.
We use matched controls with the same pipeline skeleton and vary only selected parts of $\Theta$.
The key control is \textit{Tuned-Clf}: SSL remains at defaults, while the same 100-trial budget searches the same classifier space as SemiScope.
This comparison estimates the residual value of joint SSL and classifier search; comparing Tuned-Clf with untuned defaults estimates classifier HPO alone.
Ablations in Section~\ref{sec:results} remove one component at a time.
Equal compute and the same validation protocol make the attribution meaningful; otherwise, gains could reflect more search rather than a better formulation.
\FloatBarrier

%% file: sections/experiment.tex
\begin{table}[!t]
\centering
\small
\setlength{\tabcolsep}{4pt}
\caption{Evaluation datasets and class imbalance.
Label $= 1$ denotes the minority (threat) class.}
\begin{tabular}{l|l|c|c|c}
\hline
\textbf{Dataset} & \textbf{Task} & \textbf{Rows} & \textbf{Features} & \textbf{Imbalance} \\ \hline \hline
CIC-IDS-2017~\cite{sharafaldin2018toward} & Intrusion detection & 14,139 & 70 & 19.2\% \\ \hline
Drebin~\cite{arp2014drebin} & Malware detection & 4,126 & 215 & 16.0\% \\ \hline
NSL-KDD~\cite{tavallaee2009detailed} & Intrusion detection & 9,779 & 122 & 18.0\% \\ \hline
Phishing~\cite{aburrous2008intelligent} & Phishing detection & 5,681 & 48 & 12.0\% \\ \hline
UNSW-NB15~\cite{moustafa2015unsw} & Traffic classification & 11,982 & 196 & 20.0\% \\ \hline
\end{tabular}
\label{tbl:dataset}
\end{table}

\section{Experiment}\label{sec:evaluation}

\subsection{Research Questions and Setup}

The experiment follows the decomposition introduced above.
\textbf{RQ1} first establishes whether joint pipeline optimization improves over default SSL, AutoML, and supervised references.
\textbf{RQ2} then tests the paper's central attribution question: how much of that improvement comes from classifier HPO alone, and how much remains for joint SSL--classifier search?
\textbf{RQ3} breaks the pipeline apart through component ablations.
\textbf{RQ4} measures label efficiency by asking when joint optimization matches fully supervised performance.
\textbf{RQ5} finally evaluates whether the resulting gains justify the optimization cost.

\subsection{Datasets and Preprocessing}\label{sec:datasets}

We evaluate on five established binary tabular security benchmarks (Table~\ref{tbl:dataset}) spanning four task types.
All have 12--20\% minority-class imbalance.
To keep experiments tractable while preserving class ratios, datasets larger than 15{,}000 rows are stratified-sampled to $\approx$15{,}000 rows using seed 42; smaller datasets retain all available rows.
For CIC-IDS-2017, we also remove duplicate network-flow records before sampling, because repeated rows are a known artifact of the raw capture.
Removing duplicate records reduces potential train--test leakage and produces a more conservative subset than reports that retain duplicates.
Therefore, CIC-IDS-2017 scores are intended for comparisons among treatments under our preprocessing pipeline, not for direct comparison with studies that use different preprocessing.
Categorical features in NSL-KDD (\texttt{protocol\_type}, \texttt{service}, \texttt{flag}) and UNSW-NB15 (\texttt{proto}, \texttt{service}, \texttt{state}) are one-hot encoded.
The label column is binarized (0 = benign, 1 = threat) and moved last.
Drebin features are already binary API-call indicators, and no additional feature selection is applied.
Because Drebin access is controlled by the upstream provider, we do not redistribute the raw Drebin data.

For each seed, we first split every dataset into training (64\%), validation (16\%), and test (20\%) partitions by stratified sampling.
We then split the training partition into labeled ($r$\%) and unlabeled ($1{-}r$\%) pools for label rates $r \in \{10, 20, \ldots, 90\}$\%.
Because the paper's main decomposition is evaluated at the most label-scarce setting, the 10\% label rate receives ten seeds for Tables~\ref{tbl:rq1},
\ref{tbl:stats}, \ref{tbl:secondary}, and~\ref{tbl:ablation}.
The remaining 20--90\% label rates receive three seeds for the label-rate curves in Figure~\ref{fig:curves}.
We report mean $\pm$ standard deviation throughout.

\subsection{Baselines}\label{sec:baselines}

The baseline set is designed to separate performance comparison from causal attribution.
SemiScope is compared against semi-supervised baselines and supervised references.
Except for the exploratory AutoML baseline noted below, all treatments use the same retrain, decision-threshold tuning, and single-shot test protocol (Section~\ref{sec:framework}).

\textit{Supervised references} anchor the upper end of the comparison under full labeling.
RF, XGBoost, and LightGBM are trained with 100\% labels, each in default and 100-trial Optuna-tuned forms.
These rows are targets rather than direct competitors.

\textit{Default SSL baselines} then show what standard SSL pipelines achieve before any pipeline tuning.
Default LP/LS/ST + RF use scikit-learn's default method hyperparameters paired with a default-hyperparameter Random Forest.
LP/LS use a 7-nearest-neighbor kernel, $\alpha = 0.2$ for LS, and 1000 maximum iterations.
ST uses an RF base estimator, internal acceptance threshold $0.75$, $k_{\text{best}} = 10$, and 10 self-training iterations.
LP/LS + SMOTE apply SMOTE to the pseudo-labeled pool ($k=5$, target minority fraction $\rho{=}0.5$).

\textit{Tuned-Clf baselines} isolate the value of classifier tuning on top of those default SSL outputs.
LP+HPO and ST+HPO keep SSL at the defaults above and spend the full 100 Optuna trials searching over the \textit{same} classifier family and hyperparameter space as SemiScope.
These are the key decomposition controls.

\textit{AutoML} provides an exploratory comparison with a general-purpose tuner.
FLAML~\cite{wang2021flaml} runs on pseudo-labeled data from default LP with a 60-second budget.
Because this budget differs from the 100-trial Optuna budget, we treat FLAML as illustrative, not head-to-head.

\subsection{Metrics, Statistics, and Implementation}\label{sec:metrics}

Our primary metric is \textbf{g-measure}~\cite{shu2022reducing}, the harmonic mean of recall ($pd$) and specificity ($1-pf$).
It is computed as $g = 2 \cdot pd \cdot (100-pf) / (pd + 100 - pf)$ on a 0--100 scale.
We use g-measure instead of accuracy because accuracy can be dominated by the majority class under imbalance.
We also prefer it over F-measure, which is threshold-unstable, and AUC-ROC, which is threshold-independent.
To separate threshold effects from ranking quality, we also report recall, FPR, F-measure, and AUC-ROC.
Decision thresholds are tuned on $D_{\text{val}}$ by grid search over $[0.05, 0.95]$ (step 0.01) for every treatment.
For 10\% labels, pairwise comparisons are paired by seed ($n = 10$) and use Wilcoxon signed-rank tests; we report Cliff's~$\delta$.

%% file: sections/results.tex
\section{Evaluation Results}\label{sec:results}

\subsection{RQ1 (Effectiveness): Comparison Against Baselines}

\begin{figure}[!t]
\centering
\includegraphics[width=\linewidth]{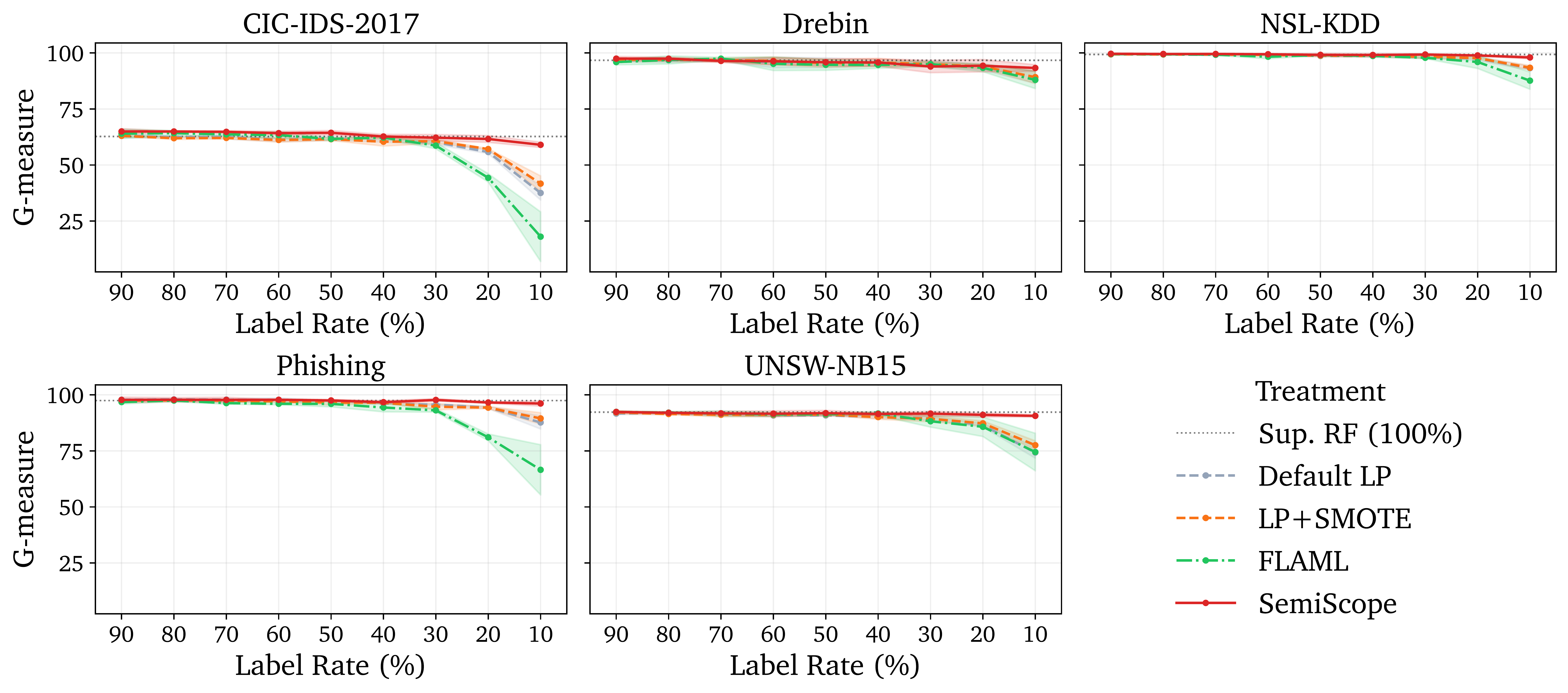}
\caption{G-measure across label rates.
We use 3~seeds at 20--90\% and 10~seeds at 10\%.
Dashed line: supervised RF (100\%).
Shaded bands: $\pm$1 std.}
\label{fig:curves}
\end{figure}

RQ1 asks whether SemiScope is consistently effective, not only whether it wins at a single operating point.
We therefore evaluate effectiveness from three complementary views.
First, Table~\ref{tbl:rq1} compares all treatments at the most label-scarce setting, i.e., 10\% labels, using 10 seeds and including the equal-budget baselines.
Second, Figure~\ref{fig:curves} traces performance across label rates from 90\% to 10\%, while the 10\% point uses the same 10 seeds, and the remaining label rates use 3 seeds.
Third, Table~\ref{tbl:stats} reports paired per-dataset statistical tests at 10\% labels.

\begin{table}[!t]
\centering
\small
\setlength{\tabcolsep}{4pt}
\caption{G-measure comparison at 10\% labels (mean $\pm$ std, 10 seeds).
\textbf{Bold}: highest mean among semi-supervised treatments per dataset.
\underline{Underline}: strongest default baseline per dataset.
$\dagger$Same 100-trial budget as SemiScope.}
\resizebox{\linewidth}{!}{%
\begin{tabular}{@{}l@{\,}!{\color{black}\vrule width 0.6pt}@{\,}rrrrr@{}}
\hline
\textbf{Treatment} & \textbf{CIC-IDS-2017} & \textbf{Drebin} & \textbf{NSL-KDD} & \textbf{Phishing} & \textbf{UNSW-NB15} \\ \hline \hline
Supervised RF (100\%) & 62.7$\pm$1.1 & 96.7$\pm$1.0 & 99.3$\pm$0.3 & 97.4$\pm$1.2 & 92.3$\pm$0.5 \\
Supervised XGBoost (100\%) & 64.7$\pm$0.7 & 97.1$\pm$0.8 & 99.5$\pm$0.2 & 97.5$\pm$1.0 & 92.5$\pm$0.8 \\
Supervised LightGBM (100\%) & 65.0$\pm$0.6 & 96.9$\pm$1.0 & 99.5$\pm$0.2 & 97.8$\pm$1.0 & 92.9$\pm$0.7 \\
Tuned Sup.\ RF$^\dagger$ (100\%) & 62.9$\pm$0.9 & 96.4$\pm$0.6 & 99.3$\pm$0.3 & 97.7$\pm$0.8 & 92.0$\pm$0.8 \\
Tuned Sup.\ XGBoost$^\dagger$ (100\%) & 64.2$\pm$1.3 & 96.9$\pm$0.9 & 99.3$\pm$0.2 & 97.0$\pm$2.0 & 92.8$\pm$0.5 \\
Tuned Sup.\ LightGBM$^\dagger$ (100\%) & 64.9$\pm$0.8 & 97.2$\pm$1.1 & 99.5$\pm$0.2 & 98.0$\pm$1.0 & 92.9$\pm$0.6 \\ \hline
Default LP + RF & 37.5$\pm$3.2 & 89.2$\pm$2.8 & 93.2$\pm$0.9 & 87.7$\pm$2.9 & 74.3$\pm$2.6 \\
Default LS + RF & 37.5$\pm$3.2 & 87.1$\pm$3.5 & 93.3$\pm$1.3 & 87.9$\pm$2.0 & 74.5$\pm$2.3 \\
Default ST + RF & \underline{46.3$\pm$2.2} & \underline{91.0$\pm$1.6} & \underline{97.2$\pm$0.8} & \underline{92.8$\pm$2.9} & \underline{85.4$\pm$1.6} \\
LP + SMOTE & 41.7$\pm$3.5 & 89.2$\pm$2.8 & 93.4$\pm$1.0 & 89.5$\pm$2.5 & 77.5$\pm$2.0 \\
LS + SMOTE & 41.7$\pm$3.5 & 86.8$\pm$3.5 & 93.5$\pm$1.2 & 88.0$\pm$1.6 & 77.3$\pm$1.9 \\
FLAML (AutoML) & 18.0$\pm$11.1 & 88.0$\pm$3.9 & 87.6$\pm$3.9 & 66.6$\pm$11.2 & 74.5$\pm$8.4 \\ \hline
Tuned-Clf: LP+HPO$^\dagger$ & \textbf{59.1$\pm$1.2} & 93.0$\pm$1.8 & 97.9$\pm$0.8 & 95.5$\pm$1.1 & \textbf{90.6$\pm$0.9} \\
Tuned-Clf: ST+HPO$^\dagger$ & 58.3$\pm$0.7 & 93.0$\pm$1.9 & 97.7$\pm$0.8 & 95.5$\pm$1.8 & 90.3$\pm$0.8 \\ \hline
SemiScope & 59.0$\pm$1.2 & \textbf{93.3$\pm$1.5} & \textbf{98.0$\pm$0.7} & \textbf{96.1$\pm$1.2} & 90.6$\pm$0.6 \\
\hline
\end{tabular}%
}
\label{tbl:rq1}
\end{table}

At the 10\% label rate, RQ1 is answered clearly against the default baselines:
SemiScope outperforms every default SSL treatment on all five datasets.
Relative to the strongest default baseline, the improvement ranges from 0.7~g-measure points on NSL-KDD to 12.7 on CIC-IDS-2017.
After Holm--Bonferroni correction, all five Wilcoxon tests remain significant (adjusted $p \leq .020$), with Cliff's $\delta \geq 0.56$.
These results establish SemiScope as a strong default-baseline competitor.
Because Tuned-Clf uses the same search budget and remains close to SemiScope, these RQ1 results motivate the attribution analysis in RQ2.

The aggregate comparison in Table~\ref{tbl:rq1} also shows why no single default baseline is sufficient.
First, default LP behaves very differently across datasets at the 10\% label rate:
it drops to 37.5 on CIC-IDS-2017 ($-$25.2 vs.\ supervised RF) and 74.3 on UNSW-NB15, but still reaches 93.2 on NSL-KDD.
This spread means that conclusions based on one dataset or one default SSL method would be fragile.
SemiScope reduces the largest LP gaps, improving over Default LP by 21.5~g-measure points on CIC-IDS-2017 and 16.3 on UNSW-NB15.

Second, adding SMOTE to default graph-based SSL has mixed effects.
LP + SMOTE improves over LP by 4.2 on CIC-IDS-2017 and 3.3 on UNSW-NB15,
but its effect is near-zero on Drebin and NSL-KDD.
This pattern is consistent with the risk described earlier: oversampling amplifies the distribution produced by SSL, so noisy pseudo-labels can limit or even reverse the benefit of synthetic samples.
SemiScope addresses this interaction by letting the optimizer decide both whether and how aggressively to oversample.

Third, Self-Training is the strongest default baseline.
When paired with decision-threshold tuning, default ST achieves 46.3--97.2 g-measure.
On NSL-KDD, it nearly matches SemiScope: 97.2 versus 98.0 in g-measure.
On the harder datasets, however, SemiScope's mean remains higher by +3.3 to +12.7.

Finally, FLAML is unreliable at low label rates in this setup.
It has high variance ($\pm$8--11 on three datasets) and collapses on Phishing (66.6$\pm$11.2).
One plausible explanation is that FLAML optimizes the classifier while treating noisy pseudo-labels as ground truth.

The descriptive gains above are supported by paired statistical tests.
Table~\ref{tbl:stats} (top panel) compares SemiScope against Default ST + RF, the strongest \textit{default} baseline.
The tests are paired by seed ($n = 10$) and use Wilcoxon signed-rank tests plus Cliff's~$\delta$.
All five comparisons yield large effects ($\delta \geq 0.56$) and remain significant after Holm--Bonferroni correction across datasets.
The top panel therefore supports the RQ1 performance claim, and the bottom panel is included here for completeness and interpreted in RQ2.

\begin{table}[!t]
\centering
\small
\setlength{\tabcolsep}{3pt}
\caption{Paired tests at 10\% labels ($n = 10$ seeds). $\Delta$ is mean paired difference in g-measure.
Bracketed ranges are bootstrap 95\% confidence intervals (CIs) computed with 5,000 resamples.
Top panel: SemiScope vs.\ Default ST + RF (large-effect difference test).
Bottom panel: SemiScope vs.\ equal-budget Tuned-Clf: LP+HPO (paired TOST~\cite{lakens2017equivalence} at the primary SESOI $=\pm 1.0$).
Holm--Bonferroni adjustment is applied across the five datasets.
\textbf{Equiv.}: $p_{\text{TOST,adj}} < 0.05$ (Drebin $.0499$).}
\resizebox{\linewidth}{!}{%
\begin{tabular}{l|c|c|c|c|c}
\hline
\textbf{Dataset} & \textbf{$\Delta$ mean (95\% CI)} & \textbf{Cliff's $\delta$ (95\% CI)} & \textbf{$p_{\text{Wilcox,adj}}$} & \textbf{$p_{\text{TOST,adj}}$} & \textbf{Equiv.} \\ \hline \hline
\multicolumn{6}{l}{\textit{SemiScope vs.\ Default ST + RF (strongest default)}} \\ \hline
\textbf{CIC-IDS-2017} & \textbf{+12.75} & \textbf{1.00} & \textbf{.010} & --- & --- \\
\textbf{Drebin} & \textbf{+2.33} & \textbf{0.68} & \textbf{.018} & --- & --- \\
\textbf{NSL-KDD} & \textbf{+0.74} & \textbf{0.56} & \textbf{.020} & --- & --- \\
\textbf{Phishing} & \textbf{+3.30} & \textbf{0.78} & \textbf{.018} & --- & --- \\
\textbf{UNSW-NB15} & \textbf{+5.20} & \textbf{1.00} & \textbf{.010} & --- & --- \\ \hline
\multicolumn{6}{l}{\textit{SemiScope vs.\ Tuned-Clf: LP+HPO (equal-budget equivalence test, SESOI $=\pm 1.0$)}} \\ \hline
CIC-IDS-2017 & $-$0.07 \,[$-$0.75,\,+0.53] & 0.00 \,[$-$0.52,\,+0.54] & 1.000 & \textbf{.040} & \textbf{\checkmark} \\
Drebin & +0.27 \,[$-$0.36,\,+0.82] & 0.10 \,[$-$0.42,\,+0.62] & 1.000 & \textbf{.0499} & \textbf{\checkmark} \\
NSL-KDD & +0.10 \,[$-$0.16,\,+0.39] & 0.07 \,[$-$0.45,\,+0.59] & 1.000 & \textbf{$<$.001} & \textbf{\checkmark} \\
Phishing & +0.64 \,[$+$0.09,\,+1.26] & 0.29 \,[$-$0.26,\,+0.76] & .420 & .137 & \ding{55} \\
UNSW-NB15 & +0.02 \,[$-$0.55,\,+0.50] & 0.02 \,[$-$0.52,\,+0.56] & 1.000 & \textbf{.014} & \textbf{\checkmark} \\
\hline
\end{tabular}%
}
\label{tbl:stats}
\end{table}

Because g-measure combines recall and specificity, we also check whether the gains reflect broader metric improvements.
Table~\ref{tbl:secondary} reports recall,
FPR, AUC-ROC, and F-measure for SemiScope and Default ST + RF at 10\%.
SemiScope achieves higher mean AUC-ROC, recall, and F-measure on all five datasets.
The main tradeoff appears on CIC-IDS-2017, where SemiScope accepts a higher false-alarm rate in exchange for substantially higher recall (+26.3).
UNSW-NB15 shows the same tradeoff more mildly: recall rises by 13.3 points, while FPR also increases.

\begin{table}[!t]
\centering
\small
\setlength{\tabcolsep}{3pt}
\caption{Secondary metrics at 10\% labels.
Default ST + RF is the strongest default baseline.
Mean over 10 seeds; better value in each pair \textbf{bolded}. Arrows indicate direction.}
\begin{tabular}{l|cccc|cccc}
\hline
 & \multicolumn{4}{c|}{\textbf{SemiScope}} & \multicolumn{4}{c}{\textbf{Default ST + RF}} \\
\textbf{Dataset} & Rec.$\uparrow$ & FPR$\downarrow$ & AUC$\uparrow$ & F1$\uparrow$ & Rec.$\uparrow$ & FPR$\downarrow$ & AUC$\uparrow$ & F1$\uparrow$ \\ \hline
CIC-IDS-2017 & \textbf{57.9} & 39.4 & \textbf{64.8} & \textbf{35.8} & 31.6 & \textbf{13.2} & 61.9 & 33.8 \\
Drebin & \textbf{90.9} & \textbf{4.1} & \textbf{98.1} & \textbf{85.7} & 87.4 & 5.1 & 96.4 & 81.7 \\
NSL-KDD & \textbf{96.9} & \textbf{0.9} & \textbf{99.7} & \textbf{96.5} & 95.6 & 1.0 & 98.8 & 95.5 \\
Phishing & \textbf{95.2} & \textbf{2.9} & \textbf{99.2} & \textbf{88.1} & 90.7 & 4.6 & 98.1 & 80.8 \\
UNSW-NB15 & \textbf{91.0} & 9.7 & \textbf{97.1} & \textbf{79.3} & 77.7 & \textbf{5.1} & 92.8 & 78.4 \\
\hline
\end{tabular}
\label{tbl:secondary}
\end{table}

\begin{finding}
\textbf{Finding 1 (Effectiveness).} Joint pipeline optimization exceeds every default baseline on all five datasets at 10\% labels.
Gains over the strongest default (Default ST + RF) range from +0.7 to +12.7 and remain significant after Holm--Bonferroni correction (Table~\ref{tbl:stats}).
\end{finding}

\subsection{RQ2 (Decomposition): Classifier HPO vs.\ Joint Search}\label{sec:rq2}

RQ1 shows that SemiScope improves over default SSL baselines, but that comparison does not tell us which part of the pipeline creates the gain.
RQ2 therefore asks whether the improvement requires \textit{joint} SSL and classifier search, or whether it mostly comes from tuning the classifier on default SSL output.
The two Tuned-Clf baselines in Table~\ref{tbl:rq1} provide this attribution control:
they keep SSL at defaults and use the same 100 Optuna trials, classifier search space, and validation-tuned threshold as SemiScope.
We use Tuned-Clf LP+HPO for the formal equivalence test in Table~\ref{tbl:stats}, and Tuned-Clf ST+HPO for the recovered-share calculation because Default ST + RF is the strongest default baseline.

Standard difference tests would only ask whether a nonzero gap is detectable.
RQ2 instead asks whether the remaining gap is small enough to ignore in practice.
We therefore use equivalence testing with a primary Smallest Effect Size of Interest (SESOI) of $\pm 1.0$~g-measure.
For context, this margin is on the order of the seed-to-seed variation in Table~\ref{tbl:rq1}, where standard deviations are often around one point, and it is small on the 0--100 g-measure scale.
We therefore treat a gap within $\pm 1$ point as practically negligible.

We fix the test protocol before inspecting outcomes.
For each dataset, the primary comparison is SemiScope versus Tuned-Clf: LP+HPO, evaluated with paired TOST~\cite{lakens2017equivalence}.
We apply Holm--Bonferroni correction across the five datasets.
Table~\ref{tbl:stats} reports the paired mean difference $\Delta$, bootstrap 95\% confidence intervals (CIs) computed with 5,000 resamples, and Cliff's~$\delta$ as a descriptive effect size.
The equivalence decision itself comes from the paired TOST on $\Delta$.

At the primary SESOI of $\pm 1.0$~g-measure, Tuned-Clf is statistically equivalent to SemiScope on four of the five datasets.
CIC-IDS-2017, NSL-KDD, UNSW-NB15, and Drebin pass after Holm correction, with Drebin at the boundary.
Their paired 95\% CIs lie within the $[-1, +1]$ practical-equivalence band (Table~\ref{tbl:stats}, bottom panel).
Phishing is the only inconclusive dataset.
SemiScope is higher by $+0.64$~g-measure, but the CI upper bound crosses $+1.0$.
We therefore do not claim either equivalence or a reliable practical advantage on Phishing at this sample size.
Thus, under the primary margin, the full joint search does not show a practical advantage over classifier-only tuning on four datasets; the fifth remains unresolved.

We also check whether this conclusion is sensitive to the practical margin.
Table~\ref{tbl:sesoi} reports equivalence verdicts at $\pm 0.5$, $\pm 1.0$, and $\pm 1.5$.
At $\pm 0.5$, no dataset reaches equivalence with $n{=}10$, and at $\pm 1.5$, all five datasets do.
The primary $\pm 1.0$ margin is therefore the smallest tested margin that separates datasets at this sample size, so we use it as the main interpretation rather than the more permissive $\pm 1.5$ margin.

\begin{table}[!t]
\centering
\small
\caption{Sensitivity of equivalence verdicts to the Smallest Effect Size of Interest (SESOI), in g-measure points, for SemiScope versus Tuned-Clf: LP+HPO.
Verdicts use Holm-adjusted paired TOST across five datasets.
\ding{51}~=~equivalent, \ding{55}~=~inconclusive.}
\label{tbl:sesoi}
\begin{tabular}{l|ccccc|c}
\hline
\textbf{SESOI} & \textbf{CIC-IDS-2017} & \textbf{Drebin} & \textbf{NSL-KDD} & \textbf{Phishing} & \textbf{UNSW-NB15} & \textbf{Total} \\ \hline\hline
$\pm 0.5$ & \ding{55} & \ding{55} & \ding{55} & \ding{55} & \ding{55} & 0/5 \\
$\pm 1.0$ & \ding{51} & \ding{51} & \ding{51} & \ding{55} & \ding{51} & 4/5 \\
$\pm 1.5$ & \ding{51} & \ding{51} & \ding{51} & \ding{51} & \ding{51} & 5/5 \\ \hline
\end{tabular}
\end{table}

\begin{table}[!t]
\centering
\small
\caption{Component ablation for RQ3 at 10\% labels (g-measure, mean $\pm$ std over 10~seeds).
Rows remove one component from full SemiScope.}
\begin{tabular}{l|rrrrr}
\hline
\textbf{Config} & \textbf{CIC-IDS-2017} & \textbf{Drebin} & \textbf{NSL-KDD} & \textbf{Phishing} & \textbf{UNSW-NB15} \\ \hline \hline
Full SemiScope & 59.0$\pm$1.2 & 93.3$\pm$1.5 & 98.0$\pm$0.7 & 96.1$\pm$1.2 & 90.6$\pm$0.6 \\
$-$ Conf.\ filter & 59.2$\pm$1.2 & 93.6$\pm$1.8 & 98.1$\pm$0.8 & 95.8$\pm$1.4 & 90.8$\pm$0.9 \\
$-$ SSL selection & 58.8$\pm$2.3 & 93.4$\pm$1.6 & 98.0$\pm$0.7 & 95.8$\pm$0.8 & 90.7$\pm$0.7 \\
$-$ Clf.\ selection & 58.4$\pm$1.9 & 93.8$\pm$1.5 & 98.0$\pm$0.7 & 95.8$\pm$1.3 & 89.9$\pm$0.8 \\
\hline
\end{tabular}
\label{tbl:ablation}
\end{table}

These equivalence results change the interpretation of the RQ1 gains:
most of the $+0.7$ to $+12.7$ improvement over defaults is not unique to the full joint search.
To quantify how much of SemiScope's default-baseline gain is recovered by classifier tuning, let $G(M)$ denote the mean g-measure of treatment $M$ at 10\% labels.
We define recovered share as
\begin{equation}
\text{Recovered Share} =
\frac{
    G(\textrm{Tuned-Clf: ST+HPO}) - G(\textrm{Default ST + RF})
}{
    G(\textrm{SemiScope}) - G(\textrm{Default ST + RF})
}.
\end{equation}
Tuned-Clf: ST+HPO is the headline comparator for this ratio because it starts from the strongest default SSL family, and the LP+HPO variant is a robustness check.

Using Tuned-Clf: ST+HPO, classifier HPO with the same budget recovers a median 86\% of SemiScope's gain over Default ST + RF, with per-dataset shares from 66\% to 94\%.
Using Tuned-Clf: LP+HPO gives a similar median of 88\%, and four of five datasets exceed 86\%.
The primary equivalence test also shows that the residual difference between classifier-only tuning and full joint search stays within the $\pm 1.0$ band on 4 out of 5 datasets.
This shows why untuned defaults are not enough as controls, i.e., without a matched classifier-HPO baseline, gains can be attributed to SSL-side search even when classifier tuning explains most of them.

\begin{finding}
\textbf{Finding 2 (Decomposition).} Tuned-Clf LP+HPO is statistically equivalent to SemiScope on four of five datasets at SESOI~$=\pm 1.0$; Phishing remains inconclusive.
Tuned-Clf ST+HPO recovers a median 86\% of SemiScope's gain over Default ST + RF, so most of the default-baseline improvement should not be attributed to SSL-side search alone.
\end{finding}

\subsection{RQ3 (Composability): Ablation Study}

\begin{figure}[!t]
\centering
\includegraphics[width=0.78\linewidth]{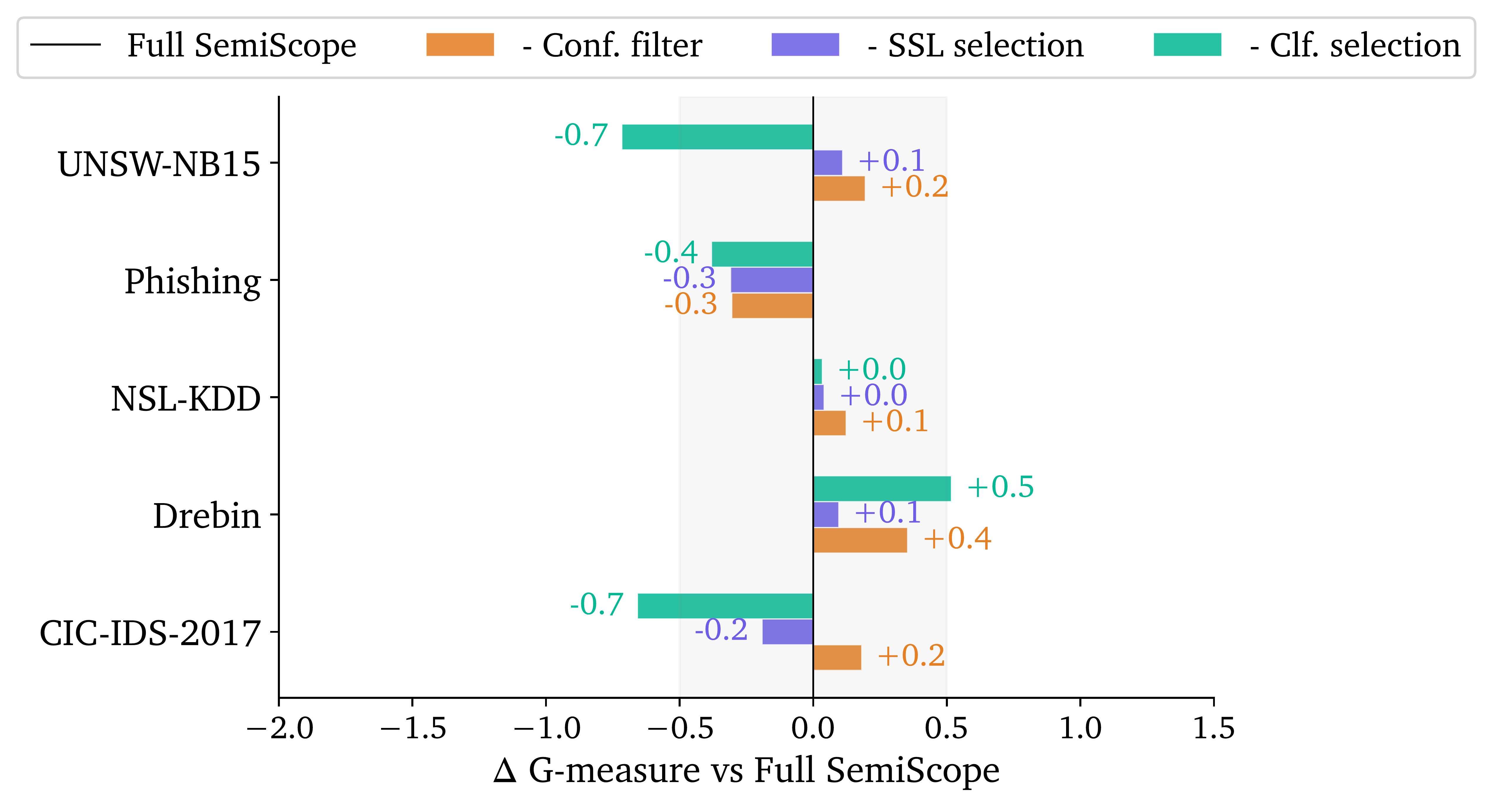}
\caption{Ablation deltas relative to full SemiScope at 10\% labels (10~seeds).
Negative values indicate lower g-measure after removing the component.}
\label{fig:ablation}
\end{figure}

RQ3 examines the same attribution question from inside the joint pipeline.
If removing one component causes most of the degradation, then that component explains more of the pipeline's value than broad interaction among all stages.
We therefore report both mean g-measure changes and changes in standard deviation, since a component can affect robustness even when its mean contribution is small.
Table~\ref{tbl:ablation} and Figure~\ref{fig:ablation} summarize the 10\% label-rate ablations.

The ablation results point to three patterns.
First, classifier selection contributes the most.
Removing it produces the largest drops on CIC-IDS-2017 and UNSW-NB15, with a $-$0.24 average change across datasets.
This is consistent with RQ2: a large part of the improvement comes from choosing and tuning the downstream classifier, especially on datasets where default SSL is weakest.

Second, SSL method selection has little effect on the average score but can affect robustness across seeds.
Fixing SSL to LP changes the mean by only $-$0.05 on average, while roughly doubling the per-seed standard deviation on CIC-IDS-2017.
Thus, the main benefit of SSL method selection in this ablation is not a consistent mean gain, but reduced sensitivity to the particular labeled/unlabeled split.

Third, confidence filtering has a small and dataset-dependent effect.
Removing it slightly improves four datasets and lowers Phishing by 0.3~g-measure, giving a near-zero average effect ($+$0.11~$\Delta$ in Table~\ref{tbl:ablation}).
This behavior matches the tunable-filter design.
When filtering does not help validation performance, the optimizer can set $t \approx 0.5$, effectively turning it off; when stricter pseudo-label selection helps, it can raise $t$ above 0.8.
This argues against imposing a single fixed confidence cutoff across datasets.

The near-neutral average effect in Table~\ref{tbl:ablation} and Figure~\ref{fig:ablation} also suggests a limitation of raw SSL confidence scores.
Because LP, LS, and Self-Training produce confidence scores on different scales, a single raw threshold on $\max_c p(c|x)$ is not guaranteed to separate clean from noisy pseudo-labels across methods.
This helps explain why filtering is not a consistently high-leverage component on these tabular datasets.
Calibrating SSL confidence scores is therefore a concrete direction for future work.
For the present study, the practical takeaway is narrower:
start with default ST, then add classifier HPO and decision-threshold tuning before spending effort on a full joint SSL search.

\begin{finding}
\textbf{Finding 3 (Composability).} Classifier selection is the highest-leverage component ($-$0.24 mean~$\Delta$ when removed).
SSL method selection mainly affects variability, and confidence filtering is near-neutral on average ($+$0.11 mean~$\Delta$).
The ablation therefore supports prioritizing classifier choice before adding more SSL-side controls.
\end{finding}

\subsection{RQ4 (Label Efficiency): Exploratory Crossover Analysis}

RQ4 shifts from fixed-label performance to label efficiency.
We ask what label rate is needed before SemiScope approaches the fully supervised reference.
We define the \emph{supervised crossover} as the smallest label rate at which SemiScope approaches Supervised~RF trained with 100\% labels.
Operationally, this means its mean g-measure, estimated with 3 seeds, is within 1~g-measure point of the supervised reference.
The within-1 criterion avoids overinterpreting exact equality from the 3-seed estimates.
Table~\ref{tbl:crossover-rq4} applies the same rule to SemiScope and Default~ST~+~RF, so the difference estimates the label-rate advantage of the optimized pipeline.

\begin{table}[!t]
\centering
\small
\caption{Label rate needed to reach within 1~g-measure point of Supervised RF (100\% labels).
Values are 3-seed directional estimates.
$\Delta$ is SemiScope minus Default ST + RF in percentage points; negative values mean SemiScope needs fewer labels.}
\label{tbl:crossover-rq4}
\begin{tabular}{l|r|r|r}
\hline
\textbf{Dataset} & \textbf{SemiScope} & \textbf{Default ST + RF} & \textbf{$\Delta$ (pp)} \\ \hline\hline
CIC-IDS-2017    & 30\% & 50\% & $-20$ \\
Drebin     & 40\% & 50\% & $-10$ \\
NSL-KDD    & 20\% & 20\% & 0 \\
Phishing   & 20\% & 20\% & 0 \\
UNSW-NB15  & 30\% & 60\% & $-30$ \\ \hline
\end{tabular}
\end{table}

In Table~\ref{tbl:crossover-rq4}, SemiScope crosses at the same or a lower label rate than Default~ST~+~RF on every dataset.
The largest label-rate advantages occur on CIC-IDS-2017 and UNSW-NB15, the same datasets where SemiScope has the largest 10\% label-rate gains over Default~ST~+~RF.
On NSL-KDD and Phishing, Default ST + RF already reaches within 1 point of the supervised reference at 20\% labels, and SemiScope does not reduce the observed crossover rate.
On Drebin, the gap is smaller (10 pp).

Because the 20--90\% label-rate curves use 3 seeds, these crossovers should be interpreted as directional evidence rather than definitive thresholds.
The pattern follows the strength of the default baseline:
when defaults drop sharply, as on CIC-IDS-2017 and UNSW-NB15, the optimized pipeline reaches the supervised reference at a lower label rate; when defaults are already near the supervised ceiling, there is little room to move the crossover earlier.
Our design concentrates statistical power on the 10\% RQ2 comparison, leaving 3 seeds at the higher label rates.
More seeds at each label rate would be needed to estimate the crossover points precisely.

\begin{finding}
\textbf{Finding 4 (Label Efficiency).} SemiScope reaches within 1 g-measure of Supervised~RF at 20--30\% labels on four datasets and at 40\% labels on Drebin.
It crosses at the same or a lower label rate than Default~ST~+~RF on every dataset, with the largest advantages on CIC-IDS-2017 ($-20$pp) and UNSW-NB15 ($-30$pp).
\end{finding}

\subsection{RQ5 (Cost and Practical Justification)}

RQ5 asks whether the additional optimization cost is reasonable in an offline model-construction setting.
The tradeoff is runtime measured against two possible benefits.
One is better performance when only 10\% of labels are available.
The other is a lower label rate for approaching the fully supervised reference.
Table~\ref{tbl:cost} summarizes these quantities using the 10\% gain over Default ST + RF and the crossover estimates from RQ4.

\begin{table}[!t]
\centering
\small
\caption{Optimization cost and label-efficiency summary.
Runtime is per SemiScope configuration with 100 Optuna trials and 4 workers.
Crossover is the SemiScope label rate from Table~\ref{tbl:crossover-rq4}.
Labels saved is measured relative to a fully labeled training set.
Gain@10\% is SemiScope minus Default ST + RF at the 10\% label rate.}
\begin{tabular}{l|r|r|r|r}
\hline
\textbf{Dataset} & \textbf{Runtime} & \textbf{Crossover} & \textbf{Labels saved} & \textbf{Gain@10\%} \\ \hline \hline
CIC-IDS-2017 & 28 min & 30\% & 70\% & +12.7 \\
Drebin & 13 min & 40\% & 60\% & +2.3 \\
NSL-KDD & 15 min & 20\% & 80\% & +0.7 \\
Phishing & 20 min & 20\% & 80\% & +3.3 \\
UNSW-NB15 & 25 min & 30\% & 70\% & +5.2 \\
\hline
\end{tabular}
\label{tbl:cost}
\end{table}

\begin{figure}[!t]
\centering
\includegraphics[width=0.65\linewidth]{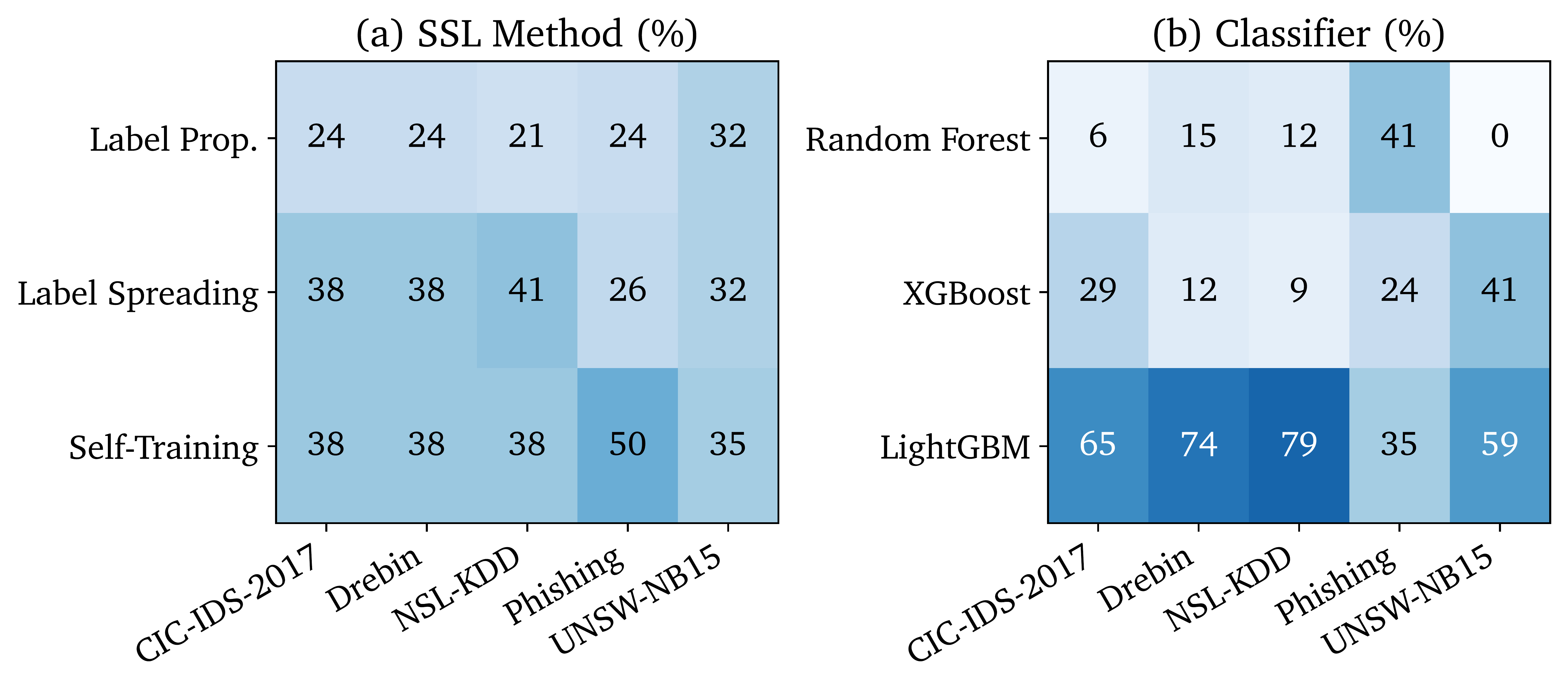}
\caption{Optimizer selections by dataset across all 170 SemiScope runs.
Darker cells are more frequent.}
\label{fig:selection}
\end{figure}

The table should be read using both benefit signals.
CIC-IDS-2017 and UNSW-NB15 provide the clearest justification because they have large gains at the 10\% label rate and also reach the supervised reference at lower label rates than Default ST + RF.
Phishing has a meaningful 10\% gain, but its crossover rate matches Default ST + RF, so its benefit is better low-label performance rather than an earlier crossover.
NSL-KDD has the weakest case because its 10\% gain is only 0.7~g-measure, even though the estimated crossover is low.
Drebin is intermediate, with a modest 10\% gain and a later crossover than the strongest cases.

The search logs give two additional signals (Figure~\ref{fig:selection}).
First, classifier choice is concentrated.
LightGBM is selected in 62\% of all SemiScope runs and is the most frequent classifier on four of the five datasets.
SSL choice is more dispersed.
Label Spreading is most frequent or tied on CIC-IDS-2017, Drebin, and NSL-KDD, while Self-Training is most frequent on Phishing and UNSW-NB15.
Second, threshold tuning is not a small adjustment.
Across the same 170 SemiScope runs, the mean tuned threshold is 0.20, and 99\% of thresholds fall below the default 0.5.
Because the minority-class prevalence is only 12--20\%, a fixed $\tau{=}0.5$ cutoff can bias predictions toward the majority class and reduce minority recall.
This is why all comparisons treat validation-set threshold tuning as part of the estimator rather than as a reporting detail.

\begin{finding}
\textbf{Finding 5 (Cost).} Optimization cost is 13--28~min per configuration.
The cost is clearest to justify on CIC-IDS-2017 and UNSW-NB15, where SemiScope has both large 10\% gains and earlier directional crossovers.
On Phishing the value is better low-label performance, while on NSL-KDD and Drebin the case for full joint search is weaker.
\end{finding}

%% file: sections/threats.tex
\section{Threats to Validity}\label{sec:threats}

\textbf{Internal Validity.} Bayesian optimization is stochastic, so we use ten seeds for the main comparison at 10\% labels and three seeds for the curves at 20--90\% labels.
This design gives the paired TOST enough power for the primary SESOI $=\pm 1.0$ on four datasets, but not for the narrower SESOI $\pm 0.5$, so we do not claim equivalence at that margin.
Independent seeds with distinct train/validation/test splits mitigate overfitting to the validation split, and the shared validation protocol should affect optimized treatments similarly.
A different hyperparameter range could shift absolute scores, but the Tuned-Clf and SemiScope comparison should remain fair because both use the same classifier space.

\textbf{External Validity.} Findings are specific to binary tabular security classification with classical SSL and classifiers based on decision trees.
The evaluation covers five benchmarks that do not include deep neural classifiers, multiclass tasks, data outside tabular form.
The claims should therefore be read as claims under fixed data splits and this protocol, not as deployment suggestions.
Within this scope, Phishing's residual advantage needs more seeds to confirm, and the curves over label rates are directional because they use only three seeds.

\textbf{Construct Validity.} SSL components are implemented with scikit-learn~\cite{pedregosa2011scikit}, Bayesian optimization uses Optuna~\cite{akiba2019optuna}, and SMOTE with a target ratio follows SMOTUNED~\cite{agrawal2018better}.
Different implementations could shift absolute numbers.
Our primary conclusion is tied to g-measure~\cite{shu2022reducing}, while alternate objectives such as F-measure could favor different tradeoffs between precision and recall.
We therefore do not claim equivalence across objectives.

%% file: sections/conclusion.tex
\section{Conclusion}\label{sec:conclusion}

In this paper, we studied semi-supervised learning for binary tabular security tasks as a configurable pipeline rather than as a single SSL algorithm.
We used SemiScope as a controlled joint-search framework and compared it with default baselines, supervised references, ablations, and equal-budget tuning controls across five security datasets.
The main finding is that joint optimization can improve SSL performance, especially when labels are scarce, but the source of that improvement is not always the SSL component itself.
Much of the gain comes from downstream model tuning and threshold tuning.
This means that reported gains from joint SSL pipelines should be interpreted through matched controls, not only through comparisons with untuned defaults.
For practitioners, we would recommend to start with a clear baseline, tune the decision threshold, and optimize the downstream classifier before investing in full joint search.
Full joint optimization should be considered when labels are expensive or default SSL leaves a performance gap.